\documentclass[10pt,twocolumn,letterpaper]{article}

\usepackage{cvpr}              

\usepackage{textcomp}
\usepackage{stfloats}
\usepackage{url}
\usepackage{verbatim}
\usepackage{graphicx}
\usepackage{algorithm}
\usepackage{multicol}
\usepackage{multirow}
\usepackage{colortbl}
\usepackage{tcolorbox}
\usepackage{bm} 
\usepackage{booktabs}
\usepackage{enumitem}
\usepackage{flushend}
\definecolor{cvprblue}{rgb}{0.21,0.49,0.74}
\usepackage[pagebackref,breaklinks,colorlinks,allcolors=cvprblue]{hyperref}

\def\paperID{7097} 
\def\confName{CVPR}
\def\confYear{2026}

\title{BiTAgent: A Task-Aware Modular Framework for Bidirectional Coupling between Multimodal Large Language Models and World Models }

\author{
Yu-Wei Zhan\textsuperscript{1},
Xin Wang\textsuperscript{1}\thanks{Corresponding authors},
Pengzhe Mao\textsuperscript{2},
Tongtong Feng\textsuperscript{1},
Ren Wang\textsuperscript{1},
Wenwu Zhu\textsuperscript{1}\footnotemark[1] \\[2mm]
\textsuperscript{1}Tsinghua University \\
\textsuperscript{2}Shandong University \\[1mm]
}

\begin{document}
\maketitle
\begin{abstract}
Building generalist embodied agents requires a unified system that can interpret multimodal goals, model environment dynamics, and execute reliable actions across diverse real-world tasks. Multimodal large language models (MLLMs) offer strong semantic priors and cross-modal generalization, while world models (WMs) provide actionable latent dynamics for prediction and control. Their combination holds promise for open-ended embodied intelligence, yet introduces two key challenges: (1) establishing a tight coupling between the semantic intent from MLLMs and the dynamic state representations within the WM’s latent space, and
(2) achieving task-aware adaptability that supports multi-task learning and cross-environment generalization. To address these limitations, we propose BiTAgent, a task-aware dynamic joint framework that enables bidirectional coupling between MLLMs and WMs. BiTAgent establishes two complementary pathways: a forward path that injects MLLM representations into the WM’s latent space for semantically guided imagination, and a backward path where WM-generated feedback refines the MLLM’s semantic space via dense text-conditioned rewards. This bidirectional interaction is realized through three synergistic components: Task-Aware Dynamic Joint Learning, Task-Aware Behavior Learning, and MLLM-WM Joint Optimization, which together harmonize semantic reasoning and dynamic prediction. Extensive experiments across multi-task and cross-environment settings demonstrate superior stability and generalization over state-of-the-art baselines, marking a step toward open-ended embodied learning.
\end{abstract}    
\section{Introduction}
\label{sec:intro}

\begin{figure}[t]
\centering
\begin{minipage}{0.98\linewidth}\centering
\centerline{\includegraphics[height=4.3cm]{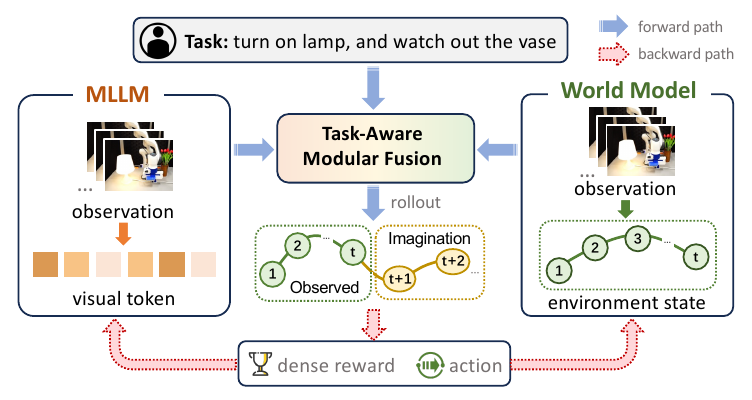}}
\end{minipage}
\vspace{-0.2cm} 
\caption{BiTAgent enables task-aware bidirectional coupling between MLLM and World Model.
In the forward path, MLLM semantics are injected into the WM via Task-Aware Modular Fusion for semantically guided imagination.
In the backward path, task-conditioned imagined trajectories produce rewards and actions, which are backpropagated through the joint loss to refine MLLM.
}\label{intro}
\vspace{-0.5cm} \medskip\end{figure}

A generalist embodied agent aims to perform diverse tasks across real-world environments within a unified framework \cite{zitkovich2023rt}. 
Such an agent is capable of interpreting high-level multimodal goals, grounding them into actionable representations, and executing reliable control based on learned environment dynamics. Instead of relying on task-specific pipelines, the ultimate goal is to build a single architecture that generalizes across heterogeneous tasks and environments while preserving consistent decision-making and control capabilities.

Recent advances in Multimodal Large Language Models (MLLMs) have demonstrated strong capabilities in semantic reasoning and cross-modal understanding \cite{radford2021clip, alayrac2022flamingo, driess2023palme, liu2023llava, openai2023gpt4}. They possess rich world knowledge and strong compositional priors, enabling them to interpret high-level task instructions and align multimodal inputs within a unified semantic space.
In parallel, World Models (WMs) excel at modeling environment dynamics and supporting decision-making \cite{ha2018worldmodels, hansen2022tdmpc, genie2024world}. They learn environment latent state representations that capture temporal dependencies and enable long-horizon imagination and planning. While MLLMs provide powerful semantic understanding and generalization across modalities, they lack physical interaction capabilities; conversely, WMs offer precise predictive dynamics and control but exhibit limited semantic abstraction and weak generalization across tasks and environments. Therefore, integrating these two paradigms offers a promising path toward open-ended embodied intelligence, where MLLMs provide semantic intent and contextual understanding, and WMs contribute physically grounded prediction and action modeling, together forming a unified framework capable of reasoning, interaction, and adaptation in diverse real-world scenarios \cite{feng2025embodied}.

Despite recent progress, existing attempts to integrate MLLMs with world models remain limited in both scope and depth. Some studies treat the MLLM as an external tool for auxiliary functions such as high-level planning \cite{song2023llm} or reward computation \cite{li2024auto, chen2024noise}.
While these methods leverage the semantic priors of MLLMs to assist in high-level decision-making and reward design, the MLLM and WM exhibit misaligned learning objectives and disconnected representations between the semantic and physical domains. A few recent studies, such as GenRL \cite{mazzaglia2024genrl} and FOUNDER \cite{wang2025founder}, attempt to address this gap by learning a connector that maps MLLM embeddings into the world model’s representation space. 
However, these approaches still face two critical limitations.
(1) They rely on one-way projection functions that transfer information solely from MLLM to the world model. The interaction with the physical environment is still handled exclusively by the world model, and no feedback from environmental dynamics is propagated back to MLLM. (2) The learned connectors are task-agnostic, applying a uniform alignment strategy across all tasks without adapting to task-specific semantics or context-dependent dynamics. The significant differences across tasks lead to task-specific parameter sensitivities, which limit the model’s capability in multitask settings and generalization across tasks. These observations motivate us to move beyond static architectural designs and explore a task-aware dynamic joint mechanism that enables bidirectional coupling between MLLMs and world models. 

To overcome these limitations, we propose BiTAgent, a \textbf{TA}sk-aware dynamic joint framework that enables \textbf{BI}directional coupling between MLLMs and world models.
As shown in Fig. \ref{intro}, the framework establishes two information pathways. In \textbf{the forward path}, the MLLM provides high-level semantic and visual representations, such as environment semantics, which are dynamically injected into the latent space of the world model through Task-Aware Fusion mechanism. This allows the world model to perform semantically guided imagination, generating trajectories that are not merely driven by physical transitions but are aligned with high-level semantic intent. 
In \textbf{the backward path}, the MLLM-WM joint model generates task-conditioned imagined trajectories and computes dense rewards that measure semantic–dynamic consistency. Since these rewards are differentiable in the latent space, , they can be backpropagated through the MLLM–WM joint loss into the MLLM, enabling its semantic representations to self-correct according to real physical dynamics.

To realize this bidirectional coupling, BiTAgent is composed of three key components: Task-Aware Dynamic Joint Learning, Task-Aware Behavior Learning, and MLLM-WM Joint Optimization. The Task-Aware Dynamic Joint Learning module integrates semantic representations from the MLLM and dynamic representations from the world model through task-conditioned modular fusion, where a gating mechanism adaptively balances the contributions of semantic and dynamic expert adapters at each layer. The Task-Aware Behavior Learning component constructs a shared imagination space for policy learning, in which rollouts are generated to align imagined trajectories with task semantics, and dense semantic consistency rewards guide physically plausible and semantically coherent behavior generation. Finally, the MLLM-WM Joint Optimization unifies semantic alignment, dynamic prediction, and behavior optimization under a single training paradigm, enabling gradient-level coupling between the MLLM and world model. In comprehensive experiments across multi-task and cross-environment generalization settings, our approach consistently outperforms state-of-the-art baselines, exhibiting strong adaptability and stability. To the best of our knowledge, this is the first work to establish a task-aware coupling framework between MLLMs and World Models, paving the way toward open-ended embodied decision making.

Our contributions are summarized as follows:
\begin{itemize}
    \item We propose BiTAgent, a task-aware dynamic joint framework that enables bidirectional coupling between MLLMs and World Models.   
    \item We introduce a Task-Aware Modular Fusion mechanism that dynamically routes information between semantic and dynamic experts under task guidance, mitigating heterogeneous task interference.
    \item We develop task-aware behavior learning with a joint optimization objective that aligns the MLLM’s semantic space with the WM’s imagination dynamics via text-conditioned dense rewards.
    \item Extensive experiments across multi-task and cross-environment settings demonstrate that BiTAgent outperforms state-of-the-art baselines, achieving superior stability and generalization in open-ended embodied learning scenarios.
\end{itemize}

\section{Related Work}

\paragraph{Multimodal Large Language Models (MLLMs).}
In recent years, the development of MLLMs has greatly advanced artificial intelligence in unified perception and reasoning. Representative models such as GPT-4o \cite{achiam2023gpt}, Gemini \cite{gemini2023}, and Claude \cite{anthropic-claude3-2024} have demonstrated strong multimodal capabilities in many multi-modal tasks. Meanwhile, open-source models such as LLaVA-Next \cite{li2024llava}, InternVL \cite{chen2024internvl}, Qwen-VL \cite{qwen2-vl-2024}, and DeepSeek-VL Janus \cite{deepseek-vl2-2024} have made multimodal research more open and reproducible. However, current MLLMs lack dynamic interacting with the real physical world, making it hard to link high-level semantics with low-level control and limiting their direct application to embodied intelligence scenarios.

\paragraph{World Models (WMs).}
World models are a key component of embodied intelligence systems. Their core objective is to learn the latent dynamics of the environment and infer the next state in either a deterministic or probabilistic manner \cite{zhang2025step}. Recent advances in world models can be broadly grouped into three paradigms. Recurrent State-Space Models (RSSMs) learn latent dynamics for state transitions, where models such as Dreamer \cite{hafner2025worldmodels} and PlaNet \cite{hafner2019learning} perform imagination-based rollouts in latent space to guide reinforcement learning. Joint Embedding Predictive Architectures (JEPAs) emphasize semantic-level consistency across states, for example, I-JEPA modeling stable and generalizable world representations from multimodal inputs \cite{assran2023self}. Generative video world models (e.g., Sora) \cite{brooks2024video} directly capture pixel-level dynamics through large-scale video generation, enabling visual prediction of future scenes. 
Despite these advances, existing world models still struggle with multi-task generalization and maintaining semantic–dynamics consistency.

\begin{figure*}[t]
\centering
\begin{minipage}{1.0\linewidth}\centering
\centerline{\includegraphics[height=7.7cm]{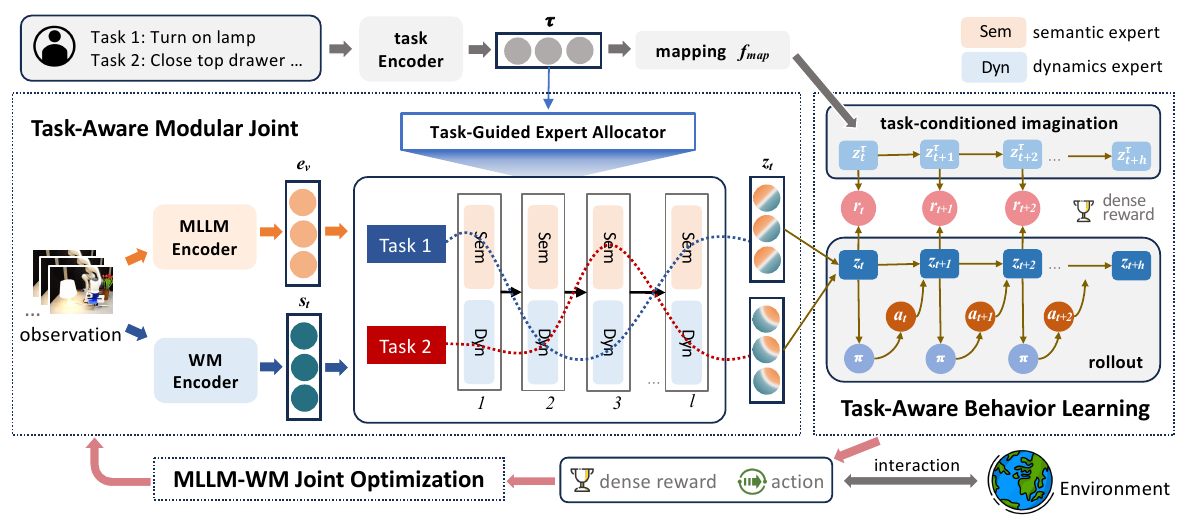}}
\end{minipage}
\vspace{-0.2cm} 
\caption{Overview of the proposed BiTAgent framework. It establishes bidirectional coupling between MLLM and World Model.
In the forward path, semantic representations produced by MLLM are injected into the WM’s latent space via the Task-Aware Modular Fusion mechanism, enabling semantically guided imagination rather than purely physics-driven rollouts.
In the backward path, Task-Aware Behavior Learning leverages WM-generated imagined trajectories to compute dense text-aligned rewards, which provide gradient feedback that reshapes the MLLM’s semantic space.}\vspace{-0.2cm} \label{framework}\medskip\end{figure*}

\paragraph{Existing MLLM–World Model Integration Paradigms.}
In recent years, MLLMs have emerged as powerful tools for task decomposition and reward specification in RL. Owing to their strong semantic understanding and reasoning capabilities, MLLMs have been adopted as planners that decompose complex long-horizon tasks into executable subgoals and interact with world models to verify task learning effectiveness. Representative works \cite{erdogan2025planandact, zhou2024dino, chiempowering} follow this paradigm, leveraging MLLMs as high-level decision modules to augment the planning capacity of world models.

In terms of reward modeling, the MLLM evaluates the performance of the world model from a visual perspective. Several works \cite{kwon2023reward, sontakke2023roboclip, klissarov2023motif, wang2024rl} compute the semantic similarity between agent states and task descriptions to generate dense reward signals from visual observations, while others \cite{3693141} leverage MLLMs to provide preference-based feedback for reward model training. These representative studies have shown that it is possible to learn effective rewards without manual design or explicit fine-tuning. However, both the planner-based and reward-based approaches treat MLLMs as external tools that extend only partial capabilities of the world models (e.g., planning or reward generation), without establishing true bidirectional communication between world interaction and MLLM.

To bridge this gap, recent works such as GenRL \cite{mazzaglia2024genrl} and FOUNDER \cite{wang2025founder} attempt to couple MLLMs and WMs within the latent state space by learning a connector that maps foundational MLLM representations into the WM’s latent dynamics, thus injecting high-level semantic priors into the world model. Nonetheless, these methods maintain a unidirectional information flow, from MLLM to WM. In contrast, our method achieves bidirectional alignment and joint optimization between MLLMs and WMs at the latent representation level, enabling embodied agents to perform unified semantic–dynamics reasoning and achieve adaptive decision-making.

\section{Methodology}

We propose BiTAgent, a \textbf{TA}sk-aware dynamic joint framework that enables \textbf{BI}directional coupling between MLLMs and world models. As illustrated in Fig. \ref{framework}, the framework is composed of three components: Task-Aware Modular Joint Learning, Task-Aware Behavior Learning, and MLLM-WM Joint Optimization. We next detail the three core modules and describe how they interact within the complete pipeline of our method.

\subsection{Task-Aware Dynamic Joint Learning}
We introduce a Dynamic Joint Architecture that seamlessly integrates the MLLM with the World Model into a unified framework. The architecture employs a task-aware modular dynamic strategy to achieve robust generalization across diverse tasks, effectively alleviating architectural conflicts in multi-task learning.

\paragraph{Dynamic Joint Architecture.}
Given a task instruction y, a video observation sequence $x_{1:t}$, and previous actions $a_{t-1}$, our framework comprises a Task Encoder, an MLLM Encoder, an RSSM-based World Model, and a Task-Aware Modular Fusion (TAMF) module:
\begin{equation}
\begin{aligned}
    Task \ Encoder: &\ \tau = f_{\text{task}}(y), \\
    MLLM \ Encoder: &\ e_v = f_\text{mllm}(x_{1:t}), \\
    World \ Model \ Encoder: & \ 
q_\phi(s_t|x_t), \\
& \ p_\theta(s_t|s_{t-1}, a_{t-1}), \\
& \ s_t = f_{\text{wm}}(x_t, s_{t-1}, a_{t-1}), \\
Modular \ Fusion : & \ 
z_t = f_{\text{mod}}(e_v, s_t, \tau), \\
\end{aligned}
\end{equation}

The task encoder transforms textual task descriptions into a global task embedding $\boldsymbol{\tau}$ through $f_{\text{task}}$, which encodes both semantic intent and task identity. This embedding serves as a dynamic routing signal for the subsequent modular layers. The MLLM encoder extracts high-level semantic representations, where $f_\text{mllm}$ encodes temporally ordered visual tokens to produce semantically consistent visual features.
The world model adopts a Recurrent State-Space Model (RSSM) architecture to capture latent dynamics in a temporally coherent form. Its encoder maps the observation $x_t$ to a latent state $s_t$ following the recurrent transition, 
$s_t = f_{\text{wm}}(x_t, s_{t-1}, a_{t-1}),$ which integrates the current observation, previous latent state, and executed action to summarize both observed environmental transitions.
The modular fusion jointly encodes semantic and dynamic representations under the guidance of $\boldsymbol{\tau}$ and produces a unified latent representation $z_t$, forming a task-conditioned feature space for imagination and behavior generation.

\paragraph{Task-Aware Modular Fusion.}
To bridge the semantic reasoning capability of the MLLM and the dynamic modeling capacity of the world model, we design a Task-Aware Modular Fusion (TAMF) module  that dynamically fuses embeddings from MLLM
and world model under task guidance. 
The module consists of L stacked layers, where each layer contains two expert adapters specialized for semantic alignment and dynamics alignment, respectively.

Given the embedding from the MLLM, $\mathbf{e}_v \in \mathbb{R}^{d_m}$,
and the latent state from the world model, $\mathbf{s}_t \in \mathbb{R}^{d_s}$,
we first compute an initial fused representation:
\begin{equation}
    \mathbf{z}^{(0)} = h_{\text{fuse}}\!\left(
    [\,\mathbf{e}_v \,;\, \mathbf{s}_t\,]
    \right),
\end{equation}
where $h_{\text{fuse}}$ is a lightweight projection layer.

Each modular layer $\ell \in \{1, \dots, L\}$ takes the previous representation $\mathbf{z}^{(\ell-1)}$ and the task embedding $\boldsymbol{\tau}$ as input.
A gating controller computes the task-conditioned routing probability,
$p^{(\ell)} = f_{gate}(\tau)$, 
which determines the contribution of each expert branch.
Then, the output of layer $l$ in the proposed TAMF is formulated as:
\begin{equation}
\mathbf{z}^{(\ell)} =
\mathbf{z}^{(\ell-1)} +
(1 - p^{(\ell)})\,\mathcal{A}^{(\ell)}_{\text{sem}}(\mathbf{z}^{(\ell-1)}) +
p^{(\ell)}\,\mathcal{A}^{(\ell)}_{\text{dyn}}(\mathbf{z}^{(\ell-1)}),
\end{equation}
where
$\mathcal{A}^{(\ell)}_{\text{sem}}$ expert focuses on semantic adaptation to align textual and visual representations;
$\mathcal{A}^{(\ell)}_{\text{dyn}}$ expert focuses on dynamics adaptation to integrate physical state transitions.

We draw inspiration from prior work on adapter-based architectures \cite{houlsby2019adapter, sung2022vladapter}, each expert adapter adopt a lightweight adaptation block structured as Pre-LayerNorm → GEGLU → Linear → LayerScale → Residual. We incorporate residual connections to mitigate training instability and ensure stable optimization.
After L layers of iterative fusion, the model produces a unified latent representation:
\begin{equation}
    \mathbf{z}^{(L)} = f_{\text{TAMF}}(\mathbf{e}_v, \mathbf{s}_t, \boldsymbol{\tau}),
    \label{tamf}
\end{equation}
which jointly encodes multimodal semantics and world dynamics, serving as a compact foundation for reconstruction, prediction, and policy optimization.

\paragraph{Task-Guided Expert Allocator.}
To enable task-aware feature routing, we introduce a lightweight gating network that transforms the task embedding $\boldsymbol{\tau}$ into a continuous gate $p \in (0,1)$, which controls the activation of semantic and dynamic experts at each layer.
Formally, given the task embedding $\boldsymbol{\tau} \in \mathbb{R}^{d_\tau}$, the gating function is defined as:
\begin{equation}
    p = \sigma\!\left(W_2\,\text{GELU}\!\left(W_1\, \text{LN}(\boldsymbol{\tau})\,\right)\right),
\end{equation}
where $\text{LN}(\cdot)$ denotes Layer Normalization, and $\sigma(\cdot)$ is the sigmoid function.
The output $p$ serves as a selection coefficient between the semantic adapter $\mathcal{A}_{\text{sem}}$ and the dynamics adapter $\mathcal{A}_{\text{dyn}}$, allowing the model to modulate fusion behavior according to the task semantics. For different tasks, the relative weighting between the semantic and dynamics branches is adaptively adjusted, allowing the model to share low-level representations across tasks while preserving task-specific expressiveness. 
Such dynamic routing mitigates gradient conflicts and interference across heterogeneous tasks, thereby enhancing both performance and generalization in multi-task scenarios.

In addition, unlike conventional single-layer multi-expert designs, the proposed Task-Aware Modular Fusion (TAMF) adopts a layer-wise dual-expert architecture, where each layer contains independent semantic and dynamics adapters, enabling localized routing and adaptive feature blending. This design brings three major benefits: first, independent gating at each layer decomposes gradient propagation into multiple local decisions, leading to clearer gradient signals and more stable convergence; second, each layer allows progressive balancing between high-level semantic abstraction and low-level physical reasoning; and third, the layer-wise routing mechanism provides inherent robustness, an inaccurate gate in one layer affects only local computation without corrupting the overall feature flow.

\subsection{Task-Aware Behavior Learning}

Building upon the fused latent representation generated by TAMF, this section focuses on how the model leverages $z_t$ to learn task-consistent behavior policies. 

\paragraph{Rollout.}
To enable the fused latent representation $\mathbf{z}_t$ to drive behavior generation, we construct a task-conditioned rollout mechanism within the shared imagination space. 
At each timestep $t$, the fused latent representation is computed according to Eq.~\ref{tamf}: 
    $\mathbf{z}_t = f_{\text{TAMF}}(\mathbf{e}_v, \mathbf{s}_t, \boldsymbol{\tau}),$
where $\mathbf{e}_v$ denotes the multimodal embedding from the MLLM, $\mathbf{s}_t$ is the latent state inferred by the RSSM-based world model $q_\phi(\mathbf{s}_t|\mathbf{x}_t)$, and $\boldsymbol{\tau}$ represents the task embedding produced by the task encoder $f_{\text{task}}(\mathcal{T})$.

Within this latent space, the imagination process unfolds according to the learned dynamics:
\begin{equation}
\begin{aligned}
    \mathbf{a}_t &\sim \pi_\psi(\mathbf{a}_t|\mathbf{z}_t), \\
    \tilde{\mathbf{z}}_{t+1} &\sim p_\theta(\mathbf{z}_{t+1}|\mathbf{z}_t, \mathbf{a}_t),
\end{aligned}
\end{equation}
where $\pi_\psi$ is the policy network and $p_\theta$ is the transition function. Iteratively applying this process for a horizon $H$ yields a sequence of imagined trajectories $\{\tilde{\mathbf{z}}_{t+h}, \mathbf{a}_{t+h}\}_{h=1}^{H}$. 
Unlike conventional model-based rollouts, our imagination process explicitly incorporates visual information from the MLLM and implicitly depends on the task embedding $\boldsymbol{\tau}$. This design enables the world model to generate task-aligned trajectories that remain semantically consistent throughout the entire temporal evolution.

\paragraph{Task-Conditioned Reward.}
In behavior modeling, the reward function plays a pivotal role in guiding the agent toward right actions.
Unlike traditional reinforcement learning where rewards are manually defined based on environmental outcomes, our framework learns dense semantic rewards directly within the latent imagination space.

Given a task embedding $\boldsymbol{\tau}$, we first map it into the latent state space of the world model through a lightweight projection:
\begin{equation}
    \mathbf{z}^{\tau}_t = f_{\text{map}}(\boldsymbol{\tau}),
\end{equation}
where $f_{\text{map}}$ ensures that the task embedding and the world model state $\mathbf{z}_t$ share the same representational manifold.
Conditioned on the current task embedding $\mathbf{z}^{\tau}_{t}$ and executed action $\mathbf{a}_t$, the text imagination module predicts the next task-aligned latent state as:
\begin{equation}
    \tilde{\mathbf{z}}_{t+1}^\tau \sim \tilde{p}_{\psi}(\mathbf{z}_{t+1}^\tau \mid \mathbf{z}^{\tau}_{t}, \mathbf{a}_t),
\end{equation}
$\tilde{\mathbf{z}}_{t+1}^\tau$ represents the expected state transition under the given task instruction.
By iteratively applying this process over a horizon $H$, we obtain a sequence of task-conditioned imagined trajectories $\{{\tilde{\mathbf{z}}_{t+h}^\tau, \mathbf{a}_{t+h}}\}_{h=1}^{H}$.

During training, we encourage the world model to align its predicted latent dynamics with those imagined from task semantics. 
This is achieved by minimizing the Kullback–Leibler divergence between the world-model rollout distribution and the task-conditioned imagination distribution:
\begin{equation}
D_{\text{KL}}\!\left(
p_{\theta}(\mathbf{z}_{t+1}\mid \mathbf{z}_t, \mathbf{a}_t)
\;\|\;
\tilde{p}_{\psi}(\tilde{\mathbf{z}}_{t+1}^\tau\mid f_{map}(\tau))
\right),\label{kl}
\end{equation}
where $p_{\theta}$ denotes the transition dynamics of the world model and $\tilde{p}_{\psi}$ represents the text-imagination model. 
This alignment enforces that the imagined trajectories are both physically consistent and semantically coherent with the given instruction.

When computing rewards for policy optimization, we measure the semantic consistency between the imagined state and the task-imagined reference as:
\begin{equation}
r_t = \text{Sim}\!\left(\mathbf{z}_t, \tilde{\mathbf{z}}_{t}^\tau\right),
\end{equation}
where $\text{Sim}(\cdot)$ denotes cosine similarity. 
This dense reward provides a smooth learning signal, guiding the policy toward behaviors that align with task semantics while maintaining physical plausibility.

\subsection{MLLM-WM Joint Optimization}

We jointly optimize the World Model and MLLM through a unified objective that enables bidirectional coupling between semantic reasoning and dynamic prediction.
The overall training objective is formulated as:
\begin{equation}
    \mathcal{L}_{\text{total}}
= \lambda_{\text{WM}}\mathcal{L}_{\text{WM}} + 
\lambda_{\text{MLLM}}\mathcal{L}_{\text{MLLM}} + 
\lambda_{\text{JBO}}\mathcal{L}_{\text{JBO}},
\end{equation}
where
$\mathcal{L}_{\text{WM}}$ focuses on reconstructing and predicting environmental dynamics, $\mathcal{L}_{\text{MLLM}}$ ensures semantic-level representation alignment, and Joint Behavior Optimization Loss $\mathcal{L}_{\text{JBO}}$ enforces task-aware joint optimization between semantic cues and dynamic rollouts.

\paragraph{World Model Loss.}
The world model learns to capture latent environment dynamics through a combination of prior--posterior consistency and observation reconstruction.
Following the recurrent state-space modeling (RSSM) formulation \cite{hafner2025worldmodels, mazzaglia2024genrl}, the overall loss is defined as:
\begin{equation}
\begin{aligned}
    \mathcal{L}_{\text{WM}}
=
\sum_{t}
\underbrace{
D_{\mathrm{KL}}\!\left[
q_{\phi}(\mathbf{z}_t \mid x_t)
\,\|\, 
p_{\theta}(\mathbf{z}_t \mid \mathbf{z}_{t-1}, a_{t-1})
\right]
}_{\text{dynamics loss}} \\
-
\underbrace{
\mathbb{E}_{q_{\phi}(\mathbf{z}_t \mid x_t)}
\!\left[
\log p_{\theta}(x_t \mid \mathbf{z}_t)
\right]
}_{\text{reconstruction loss}} .
\end{aligned}
\label{eq:wm_loss}
\end{equation}

\noindent
Here, $p_{\theta}$ and $q_{\phi}$ denote the prior and posterior latent distributions of the RSSM, respectively.
The first term enforces temporal consistency by aligning the predicted prior with the inferred posterior, 
while the second term reconstructs multimodal observations from latent states.
Together, these objectives enable the world model to learn a compact and predictive imagination space.

\paragraph{Multimodal Semantic Loss.}
The MLLM branch is trained to enhance controllable semantic representations through
semantic reconstruction and cross-modal alignment:
\begin{equation}
\mathcal{L}_{\text{MLLM}}
=
\underbrace{
\big\|\, \mathbf{e}_v - f_{dec}(z_t) \,\big\|_2^{2}
}_{\text{reconstruction loss}}
+
\underbrace{
\big\|\, \mathbf{e}_v - f_{\psi}(\tau) \,\big\|_2^{2}
}_{\text{alignment loss}},
\label{eq:mllm_loss}
\end{equation}
where $\mathbf{e}_v$ denote the visual embeddings,
$f_{\psi}$ is the representation aligner that maps linguistic features into the visual space, and $f_{dec}$ serves as the decoder that maps $\mathbf{z}_t$ back into the visual space.
This loss encourages visual reconstruction and cross-modal alignment.

\paragraph{Joint Behavior Optimization Loss.}
To align the agent’s imagined behavior with task semantics, according to Eq. \ref{kl}, we define a task-conditioned 
joint behavior optimization objective:
\begin{equation}
\mathcal{L}_{\text{JBO}}
=
-\,\mathbb{E}_{t}
\!\left[
w_{t+h} \cdot 
\text{Sim}\!\left(
\mathbf{z}_{t+h},
\tilde{\mathbf{z}}_{t+h}^{\tau}
\right)
\right],
\quad
w_{t+h} = \gamma^{t+h} ,
\label{eq:jbo_loss}
\end{equation}
where $\mathbf{z}_{t+h}$ denotes the latent trajectory generated by the world model, 
and $\tilde{\mathbf{z}}_{{t+h}}^\tau$ represents the text-conditioned imagination trajectory.
The similarity function $\text{Sim}(\cdot)$ is implemented as negative KL divergence. Follow DreamerV3 \cite{hafner2025worldmodels}, $w_{t+h} = \gamma^{t+h}$ is the discount weight. This loss bridges semantic and physical spaces, encouraging the model to generate behaviors that are both dynamically coherent 
and semantically aligned with the task prompt.

\section{Experiments}
We conduct comprehensive experiments to evaluate the effectiveness of our BitAgent.
Specifically, we aim to address three key questions:
(1) how well the model performs across multiple tasks within a single training environment,
(2) how effectively it generalizes to unseen environments when performing the same task, and
(3) how each component contributes to the overall performance improvement.

\subsection{Experimental Setup}
\paragraph{Experimental Environments.}
Our evaluation covers four locomotion control environments (Cheetah, Walker, Quadruped, and Stickman), all implemented on the DeepMind Control Suite \cite{tassa2018deepmind} frameworks.
Following GenRL, we construct offline datasets collected using the Plan2Explore strategy \cite{sekar2020planning} and replay buffers from reinforcement learning agents trained on domain-specific tasks, encompassing diverse semantic instructions and action trajectories.
Since existing work \cite{mazzaglia2024genrl} does not include the Quadruped environment, we additionally collect and preprocess it.
Detailed task definitions and task prompts are provided in the Appendix.

\paragraph{Baseline.}
We compare our proposed approach against both model-free and model-based reinforcement learning methods.
For the model-free baselines, we adopt three representative algorithms: the off-policy RL method TD3 \cite{fujimoto2018addressing}, the advantage-weighted behavior cloning method IQL \cite{kostrikov2021offline}, and the behavior-regularized approach TD3+BC \cite{fujimoto2021minimalist}. For the model-based baselines, we focus on three recent methods that also explore integrating MLLMs with World Model, such as WM-CLIP, GenRL \cite{mazzaglia2024genrl}, and FOUNDER \cite{wang2025founder}.
WM-CLIP is a variant of GenRL that learns a reversed connector, mapping latent states from the WM to the MLLM embedding space.
In contrast, GenRL and FOUNDER employ a forward connector that maps representations from the MLLM to the WM latent space, allowing semantic priors to guide imagination and planning within the world model.
Additionally, FOUNDER introduces a Temporal Distance Predictor to estimate temporally consistent reward signals during behavior learning.

\paragraph{Experimental details.}
In our experimental setup, we pretrain the world model and its associated components for 100K gradient steps, followed by another 50K updates during the behavior learning phase.
To ensure a fair comparison, our method and all model-based baselines employ the same video–language backbone, InternVideo2 \cite{wang2024internvideo2}.
The visual observations are rendered at a resolution of 64×64, with a batch size of 64 and a sequence length of 32.

\begin{table*}[t]
\centering 
\caption{Performance comparison between our method and all baselines on the DMC.
Reported scores are the mean episodic rewards over 10 random seeds (± standard error), normalized using min–max scaling where the random policy corresponds to the minimum and the expert policy to the maximum.}
\small
\vspace{-0.2cm} 
\begin{tabular}{p{70pt}|>{\hfil}p{45pt}<{\hfil}>{\hfil}p{45pt}<{\hfil}>{\hfil}p{45pt}<{\hfil}>{\hfil}p{45pt}<{\hfil}>{\hfil}p{45pt}<{\hfil}>{\hfil}p{45pt}<{\hfil}>{\hfil}p{45pt}<{\hfil}>{\hfil}p{45pt}<{\hfil}} 
\toprule
\textbf{Task} & \textbf{IQL} & \textbf{TD3+BC} & \textbf{TD3} & \textbf{WM-CLIP} & \textbf{GenRL} & \textbf{FOUNDER} & \textbf{BiTAgent} \\
\midrule
        walker stand & 0.66 ± 0.05 & 0.64 ± 0.03 & 1.01 ± 0.00 & 0.94 ± 0.01 & 1.02 ± 0.00 & 1.01 ± 0.02 & 1.03 ± 0.02 \\ 
        walker run & 0.29 ± 0.02 & 0.24 ± 0.02 & 0.35 ± 0.01 & 0.70 ± 0.01 & 0.77 ± 0.02 & 0.78 ± 0.04 & 0.87 ± 0.02 \\ 
        walker walk & 0.40 ± 0.03 & 0.44 ± 0.03 & 0.88 ± 0.02 & 0.91 ± 0.02 & 1.01 ± 0.00 & 0.94 ± 0.04 & 1.03 ± 0.01 \\ \hline
        cheetah run & 0.15 ± 0.02 & -0.01 ± 0.00 & 0.37 ± 0.01 & 0.56 ± 0.03 & 0.74 ± 0.01 & 0.81 ± 0.02 & 0.79 ± 0.02 \\ \hline
        quadruped stand & 0.52 ± 0.06 & 0.43 ± 0.05 & 0.61 ± 0.05 & 0.97 ± 0.00 & 0.97 ± 0.00 & 0.98 ± 0.01 & 1.00 ± 0.01 \\ 
        quadruped run & 0.38 ± 0.03 & 0.25 ± 0.02 & 0.26 ± 0.01 & 0.61 ± 0.02 & 0.86 ± 0.02 & 0.94 ± 0.03 & 0.95 ± 0.02 \\ 
        quadruped walk & 0.32 ± 0.02 & 0.28 ± 0.04 & 0.28 ± 0.02 & 0.92 ± 0.01 & 0.93 ± 0.01 & 0.90 ± 0.05 & 0.99 ± 0.03 \\ \hline
        stickman stand & 0.43 ± 0.04 & 0.45 ± 0.05 & 0.08 ± 0.02 & 0.32 ± 0.01 & 0.70 ± 0.02 & 0.91 ± 0.04 & 0.95 ± 0.03 \\ 
        stickman walk & 0.51 ± 0.02 & 0.46 ± 0.03 & 0.41 ± 0.02 & 0.65 ± 0.05 & 0.83 ± 0.01 & 0.91 ± 0.03 & 0.95 ± 0.03 \\ 
        stickman run & 0.23 ± 0.02 & 0.19 ± 0.02 & 0.21 ± 0.00 & 0.35 ± 0.01 & 0.35 ± 0.01 & 0.48 ± 0.02 & 0.49 ± 0.02 \\ 
        \hline
        overall & 0.39 ± 0.03 & 0.34 ± 0.03 & 0.45 ± 0.02 & 0.69 ± 0.02 & 0.82 ± 0.01 & 0.87 ± 0.03 & 0.91 ± 0.02 \\ 
        \bottomrule
    \end{tabular}\label{DMC}
\end{table*}

\subsection{Task Solving on DMC}
We evaluate our method on DMC to assess its capability in multi-task understanding and execution within a single environment.
For DMC, we select three representative locomotion tasks including Stand, Walk, and Run, and test them across multiple embodiments.

As shown in Table \ref{DMC}, model-free baselines generally perform worse.
Since they do not explicitly model environmental dynamics and instead learn policies or value functions directly from experience data, these methods struggle with long-term temporal consistency.

Compared with other model-based methods, our approach shows clear advantages. BiTAgent achieves the highest scores in 9 out of 10 tasks, demonstrating the strongest overall performance.
WM-CLIP learns a one-way mapping from world model representations to MLLM embeddings, while GenRL and FOUNDER adopt the reverse direction from MLLM to WM.
However, both directions in isolation fail to capture the complementary nature of semantic reasoning and physical dynamics.
Our joint modular fusion enables bidirectional coupling within a unified latent space, integrating the semantic abstraction capability of MLLMs with the actionable physical modeling of WMs, resulting in more accurate decision-making.

Notably, BiTAgent maintains strong performance on tasks that are already well-solved by existing methods, while delivering substantial improvements on more challenging tasks within the same environment.
For example, in the Walker domain, BiTAgent preserves competitive performance on stand and walk, while boosting the run task from 0.78 to 0.87.
These results indicate that the proposed task-aware modular fusion effectively alleviates task-specific architectural conflicts, enabling the model to retain task-unique information while preserving overall stability across diverse tasks.
\begin{figure}[t]
\centering
\begin{minipage}{0.98\linewidth}\centering
\centerline{\includegraphics[height=4.7cm]{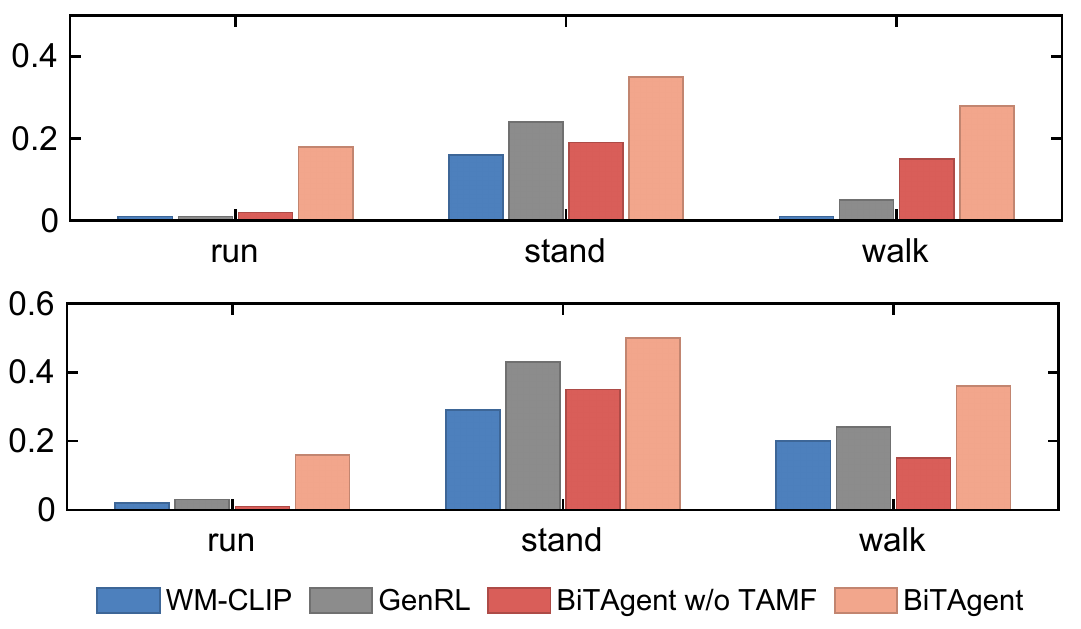}}
\end{minipage}
\vspace{-0.2cm} 
\caption{Cross-environment generalization of agents trained in the Walker domain.
The top row shows evaluation results in the Quadruped environment, while the bottom row presents results in the Stickman environment.
}\label{cross}
\vspace{-0.5cm} \medskip\end{figure}

\subsection{Cross-domain Task Solving} 
We further evaluate the cross-domain generalization capability of our method using three representative tasks: Run, Stand, and Walk. For each experiment, one environment is used as the source domain for training, and the remaining two environments serve as target domains for transfer evaluation. This configuration produces six domain combinations and eighteen task settings, enabling a comprehensive assessment. Results are shown in Fig. \ref{cross}.

Since the official implementation of FOUNDER has not been released, we compare our method with two accessible model-based baselines, WM-CLIP and GenRL. The results show that our approach achieves the best performance on all of tasks, demonstrating superior cross-domain adaptability.

These performance gains can be attributed to the TAMF module, which effectively fuses semantic and physical representations through adaptive modular routing. In addition, the modular routing mechanism enables stable sharing of low-level parameters while allowing higher-level components to adapt to task-specific requirements. As shown in the figure, the superior performance of BiTAgent over the without-TAMF variant further supports this observation.

\begin{table}[t]
    \centering
    \small
    \caption{The effectiveness of components.}\label{ab}
    \vspace{-0.2cm} 
    \begin{tabular}{p{40pt}>{\hfil}p{48pt}<{\hfil}>{\hfil}p{48pt}<{\hfil}>{\hfil}p{48pt}<{\hfil} }
    \toprule
        Method & walker stand & walker run & walker walk \\ \midrule
        $base$ & 0.89 ± 0.03 & 0.67 ± 0.01 & 0.86 ± 0.03 \\ 
        $+L_{MLLM}$ & 0.96 ± 0.02 & 0.76 ± 0.02 & 0.90 ± 0.03 \\ 
        $+TAMF$ & 1.00 ± 0.03 & 0.83 ± 0.01 & 1.00 ± 0.01 \\   \rowcolor{blue!8}$+TARO$ & 1.03 ± 0.02 & 0.87 ± 0.02 & 1.02 ± 0.01 \\
        \bottomrule
    \end{tabular}
    \vspace{-0.2cm} 
\end{table}

\subsection{Effectiveness of each Component.}

\begin{figure}[t]
\centering
\begin{minipage}{0.98\linewidth}\centering
\centerline{\includegraphics[height=9cm]{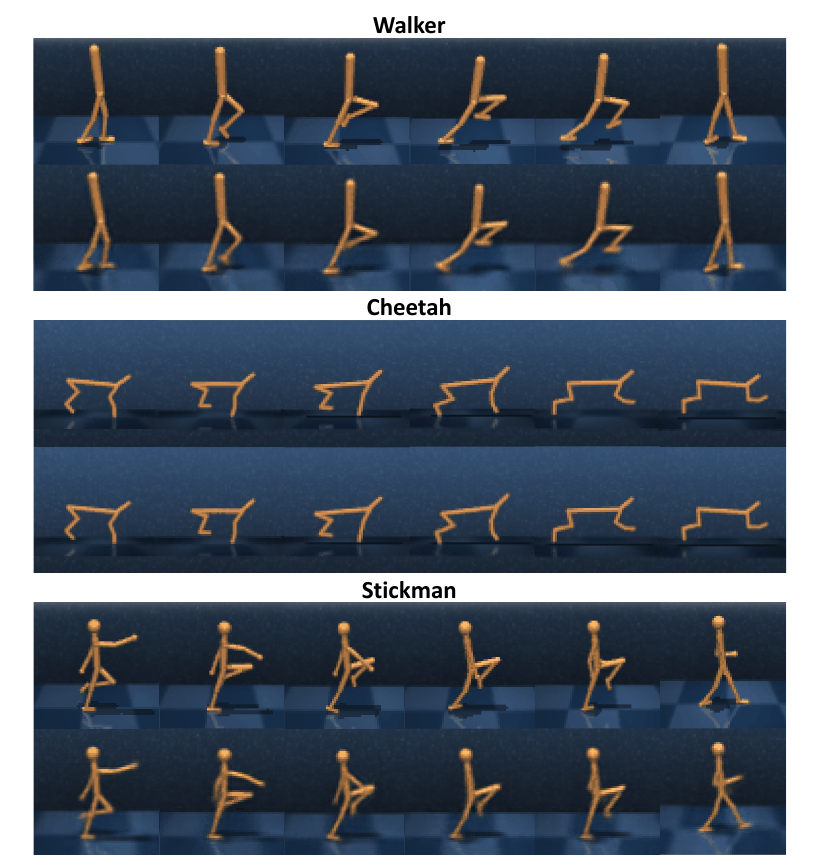}}
\end{minipage}
\vspace{-0.2cm} 
\caption{Visualization of task-conditioned imagined trajectories.
For each environment, the top row shows real observations and the bottom row shows reconstructions decoded from the task-conditioned imagination trajectories. 
}\label{visual}
\vspace{-0.5cm} \medskip\end{figure}

As shown in Table~\ref{ab}, we conduct a comprehensive ablation study to examine the contribution of each key component in our framework across three tasks in the Walker environment: Stand, Walk, and Run.

\begin{itemize}[leftmargin=1.2em]
\item Base Model.
The Base Model adopts a GenRL-style mapping from the MLLM representation to the world model’s latent space, where the fused feature is obtained by direct summation of the two embeddings. Trained solely under the world model loss $\mathcal{L}_{\text{WM}}$, it achieves average scores of 0.89, 0.67, and 0.86 on the three tasks, respectively, serving as a minimal baseline.

\item + MLLM Loss.
Introducing the multimodal reconstruction loss $\mathcal{L}_{\text{MLLM}}$ improves semantic alignment between MLLM and WM representations. The scores increase to 0.96, 0.76, and 0.90, showing that language–vision alignment benefits task performance, especially in semantically guided behaviors.

\item + Task-Aware Modular Fusion.
By adding the proposed TAMF, performance further improves to 1.00, 0.83, and 1.00. This enhancement demonstrates that dynamically routing semantic and physical features per task effectively mitigates task interference.

\item + MLLM-WM Joint Optimization.
Finally, incorporating the MLLM-WM joint optimization achieves the best results, 1.03, 0.87, and 1.02. The dense, text-conditioned reward provides stable supervision in the imagination space, allowing the model to refine its behavior generation through semantic–dynamic alignment.
\end{itemize}


\subsection{Effectiveness of behavior learning.}
To evaluate the effectiveness of task-aware behavior learning, we visualize the decoded task-conditioned imagination trajectories, as shown in Fig. \ref{visual}. For each environment, the first row displays the real observations, while the second row shows the reconstructed results from the imagined trajectories. We observe that, after behavior learning, the task-conditioned imagination trajectories closely approximate the real observations. This provides strong evidence that using these task-conditioned imagined trajectories as reference targets during reward computation is effective.
\section{Conclusions}
In this paper, we propose BiTAgent, a task-aware dynamic joint framework that enables bidirectional coupling between MLLMs and World Models.
BiTAgent establishes two complementary pathways: a forward path, where MLLM-derived visual representations are dynamically injected into the WM’s latent space, and a backward path, where WM-generated rollouts and text-conditioned rewards provide feedback to refine the MLLM’s semantic space.
To realize this coupling, BiTAgent integrates three key components: Task-Aware Dynamic Joint Learning, which adaptively fuses semantic and dynamic representations under task guidance; Task-Aware Behavior Learning, which aligns imagined trajectories with task semantics through dense text-conditioned rewards; and Joint Optimization Objectives, which unify semantic reasoning and physical prediction under a single training paradigm.
Extensive experiments across multi-task and cross-environment settings demonstrate that BiTAgent achieves superior performance compared to state-of-the-art baselines.

{
    \small
    \bibliographystyle{ieeenat_fullname}
    \bibliography{main}

@String{Computer = "{IEEE} Computer" }

@String{Chelsea = "Chelsea" }

@String{Springer = "Springer-Verlag" }

@article{achiam2023gpt,
  title={Gpt-4 technical report},
  author={Achiam, Josh and Adler, Steven and Agarwal, Sandhini and Ahmad, Lama and Akkaya, Ilge and Aleman, Florencia Leoni and Almeida, Diogo and Altenschmidt, Janko and Altman, Sam and Anadkat, Shyamal and others},
  journal={arXiv preprint arXiv:2303.08774},
  year={2023}
}

@article{gemini2023,
  title   = {Gemini: A Family of Highly Capable Multimodal Models},
  author  = {Anil, Rohan and others},
  journal = {arXiv preprint arXiv:2312.11805},
  year    = {2023},
  url     = {https://arxiv.org/abs/2312.11805}
}

@misc{anthropic-claude3-2024,
  title        = {The Claude 3 Model Family: Opus, Sonnet, Haiku (Model Card)},
  author       = {{Anthropic}},
  year         = {2024},
  note         = {Model card / technical report},
  url          = {https://www.anthropic.com/news/claude-3}
}

@article{li2024llava,
  title={Llava-next-interleave: Tackling multi-image, video, and 3d in large multimodal models},
  author={Li, Feng and Zhang, Renrui and Zhang, Hao and Zhang, Yuanhan and Li, Bo and Li, Wei and Ma, Zejun and Li, Chunyuan},
  journal={arXiv preprint arXiv:2407.07895},
  year={2024}
}

@article{qwen2-vl-2024,
  title   = {Qwen2-VL: Enhancing Vision-Language Models with Stronger Reasoning and Understanding},
  author  = {Team, Qwen},
  journal = {arXiv preprint arXiv:2407.10671},
  year    = {2024},
  url     = {https://arxiv.org/abs/2407.10671}
}

@article{deepseek-vl2-2024,
  title   = {DeepSeek-VL2: Mixture-of-Experts Vision-Language Models for Advanced Multimodal Understanding},
  author  = {DeepSeek-AI},
  journal = {arXiv preprint arXiv:2412.10302},
  year    = {2024},
  url     = {https://arxiv.org/abs/2412.10302}
}

@inproceedings{chen2024internvl,
  title={Internvl: Scaling up vision foundation models and aligning for generic visual-linguistic tasks},
  author={Chen, Zhe and Wu, Jiannan and Wang, Wenhai and Su, Weijie and Chen, Guo and Xing, Sen and Zhong, Muyan and Zhang, Qinglong and Zhu, Xizhou and Lu, Lewei and others},
  booktitle={Proceedings of the IEEE/CVF Conference on Computer Vision and Pattern Recognition},
  pages={24185--24198},
  year={2024}
}

@article{zhang2025step,
  title={A Step Toward World Models: A Survey on Robotic Manipulation},
  author={Zhang, Peng-Fei and Cheng, Ying and Sun, Xiaofan and Wang, Shijie and Zhu, Lei and Shen, Heng Tao},
  journal={arXiv preprint arXiv:2511.02097},
  year={2025}
}

@article{hafner2025worldmodels,
  title     = {Mastering diverse control tasks through world models},
  author    = {Hafner, Danijar and Pasukonis, Jurgis and Ba, Jimmy and others},
  journal   = {Nature},
  volume    = {640},
  pages     = {647--653},
  year      = {2025},
  doi       = {10.1038/s41586-025-08744-2}
}

@inproceedings{hafner2019learning,
  title={Learning latent dynamics for planning from pixels},
  author={Hafner, Danijar and Lillicrap, Timothy and Fischer, Ian and Villegas, Ruben and Ha, David and Lee, Honglak and Davidson, James},
  booktitle={Proceedings of the International Conference on Machine Learning},
  pages={2555--2565},
  year={2019}
}

@inproceedings{assran2023self,
  title={Self-supervised learning from images with a joint-embedding predictive architecture},
  author={Assran, Mahmoud and Duval, Quentin and Misra, Ishan and Bojanowski, Piotr and Vincent, Pascal and Rabbat, Michael and LeCun, Yann and Ballas, Nicolas},
  booktitle={Proceedings of the IEEE/CVF Conference on Computer Vision and Pattern Recognition},
  pages={15619--15629},
  year={2023}
}

@article{brooks2024video,
  title={Video generation models as world simulators},
  author={Brooks, Tim and Peebles, Bill and Holmes, Connor and DePue, Will and Guo, Yufei and Jing, Li and Schnurr, David and Taylor, Joe and Luhman, Troy and Luhman, Eric and others},
  journal={OpenAI Blog},
  volume={1},
  number={8},
  pages={1},
  year={2024}
}

@inproceedings{
erdogan2025planandact,
title={Plan-and-Act: Improving Planning of Agents for Long-Horizon Tasks},
author={Lutfi Eren Erdogan and Hiroki Furuta and Sehoon Kim and Nicholas Lee and Suhong Moon and Gopala Anumanchipalli and Kurt Keutzer and Amir Gholami},
booktitle={Proceedings of the International Conference on Machine Learning},
year={2025},
url={https://openreview.net/forum?id=ybA4EcMmUZ}
}

@article{zhou2024dino,
  title={Dino-wm: World models on pre-trained visual features enable zero-shot planning},
  author={Zhou, Gaoyue and Pan, Hengkai and LeCun, Yann and Pinto, Lerrel},
  journal={arXiv preprint arXiv:2411.04983},
  year={2024}
}

@inproceedings{chiempowering,
  title={Empowering World Models with Reflection for Embodied Video Prediction},
  author={Chi, Xiaowei and Fan, Chun-Kai and Zhang, Hengyuan and Qi, Xingqun and Zhang, Rongyu and Chen, Anthony and Chan, Chi-Min and Xue, Wei and Liu, Qifeng and Zhang, Shanghang and others},
  booktitle={Proceedings of the International Conference on Machine Learning}
}

@article{kwon2023reward,
  title={Reward design with language models},
  author={Kwon, Minae and Xie, Sang Michael and Bullard, Kalesha and Sadigh, Dorsa},
  journal={arXiv preprint arXiv:2303.00001},
  year={2023}
}

@article{sontakke2023roboclip,
  title={Roboclip: One demonstration is enough to learn robot policies},
  author={Sontakke, Sumedh and Zhang, Jesse and Arnold, S{\'e}b and Pertsch, Karl and B{\i}y{\i}k, Erdem and Sadigh, Dorsa and Finn, Chelsea and Itti, Laurent},
  journal={Advances in Neural Information Processing Systems},
  volume={36},
  pages={55681--55693},
  year={2023}
}

@article{klissarov2023motif,
  title={Motif: Intrinsic motivation from artificial intelligence feedback},
  author={Klissarov, Martin and D'Oro, Pierluca and Sodhani, Shagun and Raileanu, Roberta and Bacon, Pierre-Luc and Vincent, Pascal and Zhang, Amy and Henaff, Mikael},
  journal={arXiv preprint arXiv:2310.00166},
  year={2023}
}

@inproceedings{3693141,
author = {Lee, Harrison and Phatale, Samrat and Mansoor, Hassan and Mesnard, Thomas and Ferret, Johan and Lu, Kellie and Bishop, Colton and Hall, Ethan and Carbune, Victor and Rastogi, Abhinav and Prakash, Sushant},
title = {RLAIF vs. RLHF: scaling reinforcement learning from human feedback with AI feedback},
year = {2024},
booktitle = {Proceedings of the International Conference on Machine Learning},
articleno = {1071},
numpages = {28},
}

@article{wang2024rl,
  title={Rl-vlm-f: Reinforcement learning from vision language foundation model feedback},
  author={Wang, Yufei and Sun, Zhanyi and Zhang, Jesse and Xian, Zhou and Biyik, Erdem and Held, David and Erickson, Zackory},
  journal={arXiv preprint arXiv:2402.03681},
  year={2024}
}

@article{mazzaglia2024genrl,
  title={GenRL: Multimodal-foundation world models for generalization in embodied agents},
  author={Mazzaglia, Pietro and Verbelen, Tim and Dhoedt, Bart and Courville, Aaron and Rajeswar, Sai},
  journal={Advances in neural information processing systems},
  volume={37},
  pages={27529--27555},
  year={2024}
}

@inproceedings{wang2025founder,
  title={Founder: Grounding foundation models in world models for open-ended embodied decision making},
  author={Wang, Yucen and Yu, Rui and Wan, Shenghua and Gan, Le and Zhan, De-Chuan},
  booktitle={Proceedings of the International Conference on Machine Learning},
  year={2025}
}

@inproceedings{houlsby2019adapter,
  title     = {Parameter-Efficient Transfer Learning for NLP},
  author    = {Houlsby, Neil and Giurgiu, Andrei and Jastrzebski, Stanislaw and Morrone, Bryan and De Laroussilhe, Quentin and Gesmundo, Andrea and Attariyan, Mona and Gelly, Sylvain},
  booktitle = {Proceedings of the International Conference on Machine Learning},
  year      = {2019}
}

@inproceedings{sung2022vladapter,
  title     = {VL-Adapter: Parameter-Efficient Transfer Learning for Vision-and-Language Tasks},
  author    = {Sung, Yi-Lin and Nagrani, Arsha and Arnab, Anurag and Li, Shangbang and Schmid, Cordelia},
  booktitle = {Proceedings of the IEEE/CVF Conference on Computer Vision and Pattern Recognition},
  year      = {2022}
}

@inproceedings{zitkovich2023rt,
  title={Rt-2: Vision-language-action models transfer web knowledge to robotic control},
  author={Zitkovich, Brianna and Yu, Tianhe and Xu, Sichun and Xu, Peng and Xiao, Ted and Xia, Fei and Wu, Jialin and Wohlhart, Paul and Welker, Stefan and Wahid, Ayzaan and others},
  booktitle={Conference on Robot Learning},
  pages={2165--2183},
  year={2023},
  organization={PMLR}
}

@article{tassa2018deepmind,
  title={Deepmind control suite},
  author={Tassa, Yuval and Doron, Yotam and Muldal, Alistair and Erez, Tom and Li, Yazhe and Casas, Diego de Las and Budden, David and Abdolmaleki, Abbas and Merel, Josh and Lefrancq, Andrew and others},
  journal={arXiv preprint arXiv:1801.00690},
  year={2018}
}

@inproceedings{sekar2020planning,
  title={Planning to explore via self-supervised world models},
  author={Sekar, Ramanan and Rybkin, Oleh and Daniilidis, Kostas and Abbeel, Pieter and Hafner, Danijar and Pathak, Deepak},
  booktitle={Proceedings of the International Conference on Machine Learning},
  pages={8583--8592},
  year={2020},
  organization={PMLR}
}

@inproceedings{fujimoto2018addressing,
  title={Addressing function approximation error in actor-critic methods},
  author={Fujimoto, Scott and Hoof, Herke and Meger, David},
  booktitle={Proceedings of the International Conference on Machine Learning},
  pages={1587--1596},
  year={2018},
  organization={PMLR}
}

@article{kostrikov2021offline,
  title={Offline reinforcement learning with implicit q-learning},
  author={Kostrikov, Ilya and Nair, Ashvin and Levine, Sergey},
  journal={arXiv preprint arXiv:2110.06169},
  year={2021}
}

@article{fujimoto2021minimalist,
  title={A minimalist approach to offline reinforcement learning},
  author={Fujimoto, Scott and Gu, Shixiang Shane},
  journal={Advances in neural information processing systems},
  volume={34},
  pages={20132--20145},
  year={2021}
}

@inproceedings{wang2024internvideo2,
  title={Internvideo2: Scaling foundation models for multimodal video understanding},
  author={Wang, Yi and Li, Kunchang and Li, Xinhao and Yu, Jiashuo and He, Yinan and Chen, Guo and Pei, Baoqi and Zheng, Rongkun and Wang, Zun and Shi, Yansong and others},
  booktitle={Proceedings of the European Conference on Computer Vision},
  pages={396--416},
  year={2024},
  organization={Springer}
}

@inproceedings{radford2021clip,
  title={Learning Transferable Visual Models From Natural Language Supervision},
  author={Radford, Alec and Kim, Jong Wook and Hallacy, Chris and Ramesh, Aditya and Goh, Gabriel and Agarwal, Sandhini and Sastry, Girish and Askell, Amanda and Mishkin, Pamela and Clark, Jack and Krueger, Gretchen and Sutskever, Ilya},
  booktitle={Proceedings of the International Conference on Machine Learning},
  year={2021}
}

@inproceedings{alayrac2022flamingo,
  title={Flamingo: A Visual Language Model for Few-Shot Learning},
  author={Alayrac, Jean-Baptiste and Donahue, Jeff and Luc, Pauline and Miech, Antoine and Barr, Igor and Hasson, Yana and Mensch, Arthur and Millican, Katie and Reynolds, Malcolm and Ruesch, Dieter and others},
  booktitle={Advances in Neural Information Processing Systems},
  year={2022}
}

@inproceedings{driess2023palme,
  title={PaLM-E: An Embodied Multimodal Language Model},
  author={Driess, Danny and Xia, Fei and Srinivasan, Pratyusha and Yu, Tianhe and Gehring, Jonathan and Zhao, C. Karen and Chen, Xinying and Raghunathan, Parth and Andrychowicz, Marcin and Ibarz, Julian and others},
  booktitle={Proceedings of the International Conference on Machine Learning},
  year={2023}
}

@inproceedings{liu2023llava,
  title={Visual Instruction Tuning},
  author={Liu, Haotian and Li, Chunyuan and Xu, Qiang and Li, Yong Jae},
  booktitle={Advances in Neural Information Processing Systems},
  year={2023}
}

@misc{openai2023gpt4,
  title={GPT-4 Technical Report},
  author={OpenAI},
  year={2023},
  howpublished={\url{https://openai.com/research/gpt-4}}
}

@inproceedings{ha2018worldmodels,
  title={World Models},
  author={Ha, David and Schmidhuber, J{\"u}rgen},
  booktitle={Advances in Neural Information Processing Systems},
  year={2018}
}

@inproceedings{hansen2022tdmpc,
  title={Temporal Difference Model Predictive Control},
  author={Hansen, Nicklas and Suo, Wei and Laskin, Michael and Abbeel, Pieter and Kumar, Vikash},
  booktitle={Proceedings of the International Conference on Machine Learning},
  year={2022}
}

@inproceedings{genie2024world,
  title={GENIE: Generative Interactive Environments},
  author={Brooks, Tim and Xie, Saining and others},
  booktitle={Advances in Neural Information Processing Systems},
  year={2024}
}

@inproceedings{song2023llm,
  title={Llm-planner: Few-shot grounded planning for embodied agents with large language models},
  author={Song, Chan Hee and Wu, Jiaman and Washington, Clayton and Sadler, Brian M and Chao, Wei-Lun and Su, Yu},
  booktitle={Proceedings of the IEEE/CVF International Conference on Computer Vision},
  pages={2998--3009},
  year={2023}
}

@inproceedings{li2024auto,
  title={Auto mc-reward: Automated dense reward design with large language models for minecraft},
  author={Li, Hao and Yang, Xue and Wang, Zhaokai and Zhu, Xizhou and Zhou, Jie and Qiao, Yu and Wang, Xiaogang and Li, Hongsheng and Lu, Lewei and Dai, Jifeng},
  booktitle={Proceedings of the IEEE/CVF Conference on Computer Vision and Pattern Recognition},
  pages={16426--16435},
  year={2024}
}

@article{chen2024noise,
  title={Noise contrastive alignment of language models with explicit rewards},
  author={Chen, Huayu and He, Guande and Yuan, Lifan and Cui, Ganqu and Su, Hang and Zhu, Jun},
  journal={Advances in Neural Information Processing Systems},
  volume={37},
  pages={117784--117812},
  year={2024}
}

@article{feng2025embodied,
title={Embodied AI: From LLMs to World Models},
author={Feng, Tongtong and Wang, Xin and Jiang, Yu-Gang and Zhu, Wenwu},
journal={IEEE Circuits and Systems Magazine},
year={2025}
}
}


\end{document}